\documentclass[conference]{IEEEtran}
\IEEEoverridecommandlockouts
\usepackage{cite}
\usepackage{amsmath,amssymb,amsfonts}
\usepackage{algorithmic}
\usepackage{graphicx}
\usepackage{textcomp}
\usepackage{xcolor}
\usepackage[utf8]{inputenc}

\def\BibTeX{{\rm B\kern-.05em{\sc i\kern-.025em b}\kern-.08em
    T\kern-.1667em\lower.7ex\hbox{E}\kern-.125emX}}

\usepackage[english]{babel}
\graphicspath{{images/}}
\usepackage{float}

\usepackage{blindtext}
\usepackage{subfiles}

\usepackage[export]{adjustbox}
\usepackage{wrapfig}
\usepackage{booktabs}
\usepackage{subcaption}
\usepackage{svg}

\usepackage{mathtools}
\usepackage[thinc]{esdiff}

\usepackage{multirow}
\usepackage{tabularx}

\usepackage{makecell}
\makeatletter
\newcommand\notsotiny{\@setfontsize\notsotiny{6.31415}{7.1828}}
\makeatletter

\usepackage{scrextend}
\usepackage[printonlyused]{acronym}

\usepackage[colorinlistoftodos]{todonotes}

\usepackage{tikz}
\usetikzlibrary{shapes}
\usetikzlibrary{math}
\definecolor{boxcolor}{rgb}{0.95, 0.95, 0.95}
\newcommand{\mybox}[1]{\begin{tikzpicture}[baseline=(n.base)]
        \node[draw=gray,rounded rectangle,fill=boxcolor](n) {\footnotesize{#1}};
    \end{tikzpicture}}
\newcommand{\myedge}[4]{\begin{tikzpicture}[baseline=(n.base)]
        \node[draw=gray,rounded rectangle,fill=boxcolor] (a)  at (0, 0) {\footnotesize{#1}};
        \node[] (b)  at (#4, 0) {\footnotesize{#2}};
        \node[draw=gray,rounded rectangle,fill=boxcolor] (c)  at (#4*2, 0){\footnotesize{#3}};
        \draw[gray] (a)--(b);
        \draw[gray] (b)--(c);
    \end{tikzpicture}}

\usepackage[shortlabels]{enumitem}

\RequirePackage{doi}
\usepackage[numbers]{natbib}
\usepackage{hyperref}

\newcommand{\orcid}[1]{\href{https://orcid.org/#1}{\includegraphics[width=10pt]{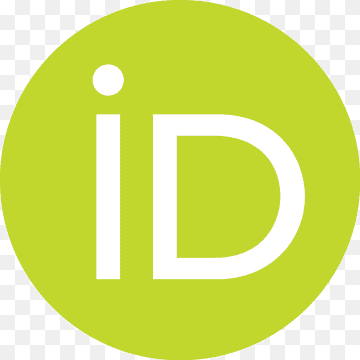}}}

\begin{document}
\title{Graph Modeling in Computer Assisted\\ Automotive Development
}

\author{\IEEEauthorblockN{1\textsuperscript{st} Anahita Pakiman\orcid{0000-0001-9706-7305}} 
\IEEEauthorblockA{
    \textit{Fraunhofer SCAI},
    \textit{Bergische Universität Wuppertal}\\
    anahita.pakiman@scai.fraunhofer.de 
}
\and
\IEEEauthorblockN{2\textsuperscript{nd} Jochen Garcke\orcid{0000-0002-8334-3695}}
\IEEEauthorblockA{\textit{Universität Bonn},
\textit{Fraunhofer SCAI}\\
jochen.garcke@scai.fraunhofer.de
}
}

\maketitle
\begin{abstract}
We consider graph modeling for a knowledge graph for vehicle development, with a focus on crash safety.
An organized schema that incorporates information from various structured and unstructured data sources is provided, which includes relevant concepts within the domain.
In particular, we propose semantics for crash computer aided engineering (CAE) data, which enables searchability, filtering, recommendation, and prediction for crash CAE data during the development process.
This graph modeling considers the CAE data in the context of the R\&D development process and vehicle safety.
Consequently, we connect CAE data to the protocols that are used to assess vehicle safety performances.
The R\&D process includes CAD engineering and safety attributes, with a focus on multidisciplinary problem-solving.
We describe previous efforts in graph modeling in comparison to our proposal, discuss its strengths and limitations, and identify areas for future work.

\end{abstract}


%
%
\section{Introduction}
\label{sec:intro}

Computer aided engineering (CAE) was introduced in the automotive industry in the 1970s, based on the experience from civil and aeronautical engineering where numerical simulations were used to study complex 
structural analysis problems.
In particular, CAE is used to study vehicle crashes, where traditionally the automotive industry relied solely on destructive physical testing of prototypes to achieve and verify crashworthiness.
In recent decades, virtual crash tests of vehicles on the computer, using commercial simulation software, supplemented the costly physical testing-only option.
Nowadays, virtual crash tests outnumber their physical counterparts by magnitudes during the development of new cars~\cite{Spethmann2009}.

Today, the Finite Element method (FEM) is the preponderant tool for automotive crash simulation~\cite{Spethmann2009}.
It consists of a discretisation of the system in space and time, where FEM calculations began with modeling the crash behavior of single components of a vehicle.
The result for a given component, or full car model, is a mesh of elements, where each element is defined by its nodes.
Model resolution, and thereby simulation accuracy, increases with the number of elements.
Nowadays, automotive crash models in some OEMs may contain up to 10 million elements, and the total number of crash simulations for a vehicle R\&D period is more than 25000.

The large amount of complex data confronts engineers with the challenge to explore the simulation results sufficiently, due to lack of engineering time and limitations of data storage, processing and analysis tools.
This need pushed the automotive companies to uptake pre- and post-processing tools to be more efficient in analysing the data, with the goal to spend the engineers time on solving the problem instead of data processing.
Nevertheless, even with all the achievements so far, data flow within the companies is still inefficient.

Yet, crash scenarios studied in the development phase are just a tiny proportion of the real crashes.
Studies show inadequate automotive safety performance to cover the diversity in anatomy characteristics~\cite{linder2018average} and in different crash scenarios.
CAE advancements in reliability led to an increased dependency of the vehicle design on the analysis configuration that performs sufficiently in regulations and rating crash tests; however, it is not safe in a broader context.
The need to increase the number of simulations and the limitation of CAE engineers' time emphasizes the importance of an intelligent system to capture domain knowledge as knowledge graphs (KGs) for automotive, which we call car-graph.

Graphs allow maintainers to postpone the definition of a schema, allowing the data – and its scope – to evolve more flexibly than typically possible in a relational setting, particularly for capturing incomplete knowledge~\cite{hogan2021knowledge}.
This data structure flexibility is an ideal fit for the automotive industry, which is a fast-evolving industry due to a short development phase of the product caused by the high market demands.
At the foundation of any KGs is the principle of first applying a graph abstraction to data, resulting in an initial data graph~\cite{hogan2021knowledge}.
Here we propose a graph data modeling for the crash analysis domain that includes the FE-model, FE-simulations, and requirement protocols\footnote
{The databases are at \href{https://github.com/Fraunhofer-SCAI/GAE-vehicle-safety}{github.com/Fraunhofer-SCAI/GAE-vehicle-safety}.
}.

A fundamental step towards making the practical utilization of car-graph a reality is transforming the data flow within companies.
In today's workflow, design engineers and attribute leaders rely on the reporting by CAE engineers.
Design engineers, or computer-aided design (CAD) engineers, are responsible for the detailed design of the vehicle, resulting in vector-based graphic representations to depict the designs in the form of surfaces of hollow solids in three-dimensional space. 
Each design engineer is typically responsible for developing several parts of the vehicle.
A complete vehicle in CAD-format is then converted to an FE-model, where each CAE engineer setups and performs a specific analysis, e.g. to study the behavior of a specific group of parts in a frontal crash.
Furthermore, an attribute leader collects all the available analysis from the CAE engineers, resolves the conflicts in design performance, and feedbacks the final decision to the CAD engineer.
Current static reporting restricts the independent exploration of the data.
The limitation arises from the company's dependency on available software and the required skills to use CAE tools.
Consequently, multi-disciplinary collaboration is ensured if the attribute leader has CAE expertise or the design engineer has a background in the discipline requirements.
Lack of multi-disciplinary collaboration degrades efficient problem-solving.

CAE data modeling is challenging since the data is complex, and several disciplines with different requirements --- CAE engineers, CAD engineers, and attribute leaders --- interact with the CAE data.
However, the flexibility of graph data modeling reflects existing uncertainties and allows the modeling to evolve.
In this work, we present an initial attempt to define a semantic representation that stores information regarding the different crash scenarios, the vehicle design deviations during the development process, and the quantities of interest that measure the outcome.
Consequently, we propose semantic selections that follow the development concepts, FE-modeling terminology, crashworthiness assessment quantities, and other relevant entities. 
Data complexity within each analysis vs. the total number of available analyses caused us to be selective in loading data, and currently, we have a relatively small graph compared to other domains.
Additionally, these can be used as input for machine learning (ML) analysis, where the graph modeling also allows storing ML results.
Our vision is to use data modeling and ML to auto-assess the cause and effect in the development process to assist engineers and, in particular, to assess the safety of different, uncalculated crash scenarios.

After covering related work in section~\ref{sec:rw}, we describe in section~\ref{sec:en_data} Euro NCAP, an example of crash test requirements~\cite{van2016european}, where a large amount of CAE data is generated during the development process to meet these requirements.
With these protocols, we look for semantics to shape the assessment structure with two targets, first benchmarking the vehicles and second supporting engineers to independently understand the CAE reports by connecting the performances to the requirements, outlined in section~\ref{sec:cae_data}.
Finally, we present the application of this data modeling, enabling semantically connected dynamic reporting with \href{https://CAEWebVis.scai.fraunhofer.de/}{CAEWebVis} and graph analytics in section~\ref{sec:applctn}.
Note that we present a first application of our graph modeling in to find analysis similarities with the Simrank++ method in a companion article~\cite{pakiman2022}.

%
%
\section{Related work}
\label{sec:rw}

The term "Knowledge Graph" was introduced in 1972 as a graph of data intended to accumulate and convey knowledge about the real world~\cite{schneider1973course}.
Currently, a KG is unavailable for the vehicle’s structural design for crash CAE analysis.
A survey in domain-specific KGs summarizes available KGs in engineering~\cite{Abu-Salih2021}, where
the most closely related engineering domain to our application is manufacturing.
However, ongoing research focuses more on production and manufacturing than product development. Example applications of KGs include digital twin models for industrial production~\cite{Banerjee2017}, industry 4.0~\cite{Garofalo2018}, and computer-aided manufacturing (CAM)~\cite{Li2018}.
However, there are investigations on ontologies for automotive industry~\cite{urbieta2021design, feld2011automotive}, FE simulation~\cite{Kestel2019a} or crash~\cite{Fatfouta2019a, Fatfouta2020}.

According to~\cite{Kestel2019a}, several studies have already applied a knowledge-based ontology system to provide simulation knowledge to FE users.
These studies disregard extracting new relationships among the data or answering analytical questions of an engineer.
In some, the focus has been on automating the generation of the FE simulation~\cite{Kestel2019a, Sun2009a} or retrieving simulation solutions from existing simulation~\cite{Wriggers2008, Kugler2018}.
However, the case studies~\cite{Kestel2019a, Wriggers2008, Sun2009a, Kugler2018} treat simpler applications in comparison to a full vehicle crash simulation.
\cite{Ziegler2021} characterized the CAE domain and identified unsolved challenges for tailored data and metadata management as a graph.
\cite{Fatfouta2019a, Fatfouta2020} has looked explicitly at a crash simulation ontology and investigated the reasoning structure of engineers, particularly regarding report generation.
Overall,~\cite{Fatfouta2019a, Fatfouta2020, Ziegler2021} have a knowledge management system orientation to understand data structure and procedures in the company, where CAE data was considered broadly.

Additionally, there is currently no universally defined strategy for building AI-oriented ontologies for the automotive sector.
One of the critical challenges is the lack of a standardized global vocabulary~\cite{urbieta2021design}.
Different works in the transportation domain have identified the importance of domain-knowledge structures for different purposes.
For example, traffic scenes real-time assessment~\cite{bagschik2018ontology}, Public Transport Systems automatic support for the design and analysis of performance monitoring ~\cite{benvenuti2017ontology}, and applications testing and labeling ~\cite{urbieta2021design}.
However, based on~\cite{urbieta2021design}, there is not yet a standardized approach or formal requirement other than the ontology languages defined by the W3C group to define an ontology.

Future vehicle developments require assessing relevant accident scenarios not addressed by today’s regulations or consumer crash tests.
As future highly automated vehicles (HAVs) are supposed to provide new, alternative sitting positions, i.e., increased travel comfort for the occupant, these positions will need safety approval and homologation.
Consequently, several ongoing projects
that define the technological developments needed to enable the automotive industry to design and develop new safety systems for advanced, safe, and comfortable sitting positions.
The OpenPASS~\cite{openPASS} software platform is an outcome of these investigations that enables the simulation of traffic scenarios to predict the real-world effectiveness of advanced driver assistance systems or automated driving functions.
Accordingly, the transportation ontology models have focused on autonomous driving and active safety.
There is a potential to assist active safety with passive safety (structural design) functionality that will lead to having a safe accident.
However, due to the scope of the work, no ontology model is defined for it to cover passive safety and the CAE domain.

%
%
\section{Euro NCAP Data Model}
\label{sec:en_data}
There are several crash safety regulations for car-makers to follow, which help to assess the safety of vehicles in the market.
These regulations cover a series of vehicle tests to represent essential real-life accident scenarios that could result in injured or killed car occupants or other road users.
The European New Car Assessment Programme (Euro NCAP) is an organization that designs and conducts such tests~\cite{van2016european}. We will use the information from Euro NCAP as a basis for part of our graph modeling.\footnote{The data is available at \href{https://www.euroncap.com/}{www.euroncap.com/}.}

\begin{figure}
    \centering
    \includegraphics[width=0.76\linewidth]{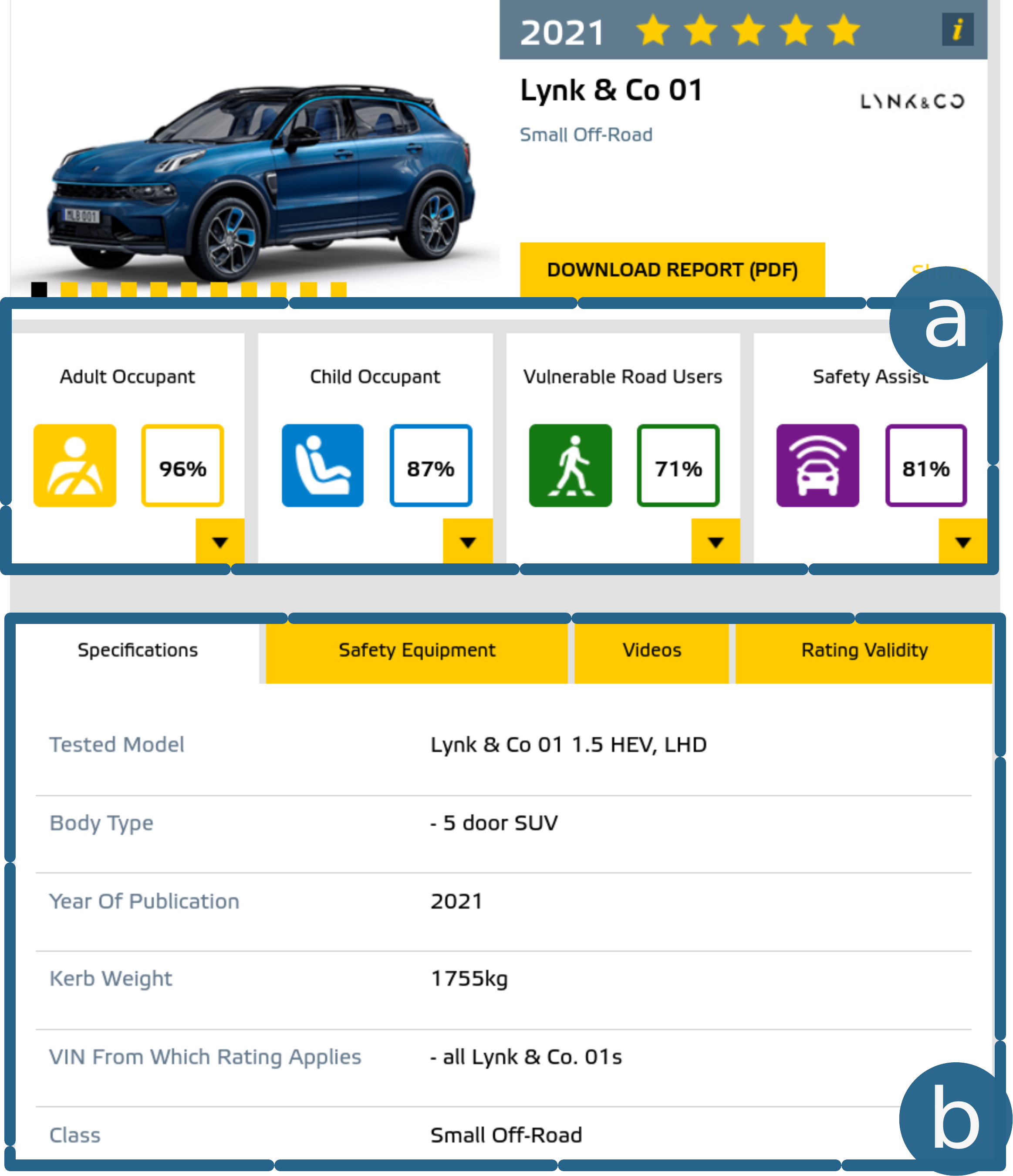}
    \caption{An example of Euro NCAP result page structure.}
     \label{fig:result_page}
\end{figure}

Euro NCAP has two primary sources of information that are of interest, test protocols and test results.
Test protocols describe the specifications and the assessment methods for each crash test, these are available as PDF documents.
Whereas a test result page contains safety performances of a released vehicle in the market as HTML pages, images, and PDF documents, see Figure~\ref{fig:result_page} for an example.
Design engineers commonly compare information about the safety performance of similar vehicle classes during a new vehicle development concept phase.
However, such an assessment requires a lot of manual data processing since more complex queries of the existing data is impossible, e.g., summarizing the pedestrian upper leg performance of vehicles with a specific ride height.
Our ultimate vision for the Euro NCAP data modeling is to add semantics to them and develop an assistance benchmarking tool for the engineers, where the raw data
can be extracted from the available test protocols and test results.

First, we look for semantics to compare the safety performance of vehicles.
Such a comparison of vehicle performance is primarily conducted with vehicles from the same class, e.g. large family cars.
We model a \mybox{Veh} node that represents a vehicle on the market.
Thus, the class specification of the vehicle is stored as a \mybox{Class} node to ease querying vehicles from the same class.
As can be seen, the available information per vehicle includes the protocol classifications for different Euro NCAP assessments, Figure~\ref{fig:result_page}(a), and the vehicle configuration, Figure~\ref{fig:result_page}(b).
We include the vehicle specification table and test media, images, and video URL as \mybox{Veh} node properties.

\begin{figure}
    \centering
    \includegraphics[width=0.9\linewidth]{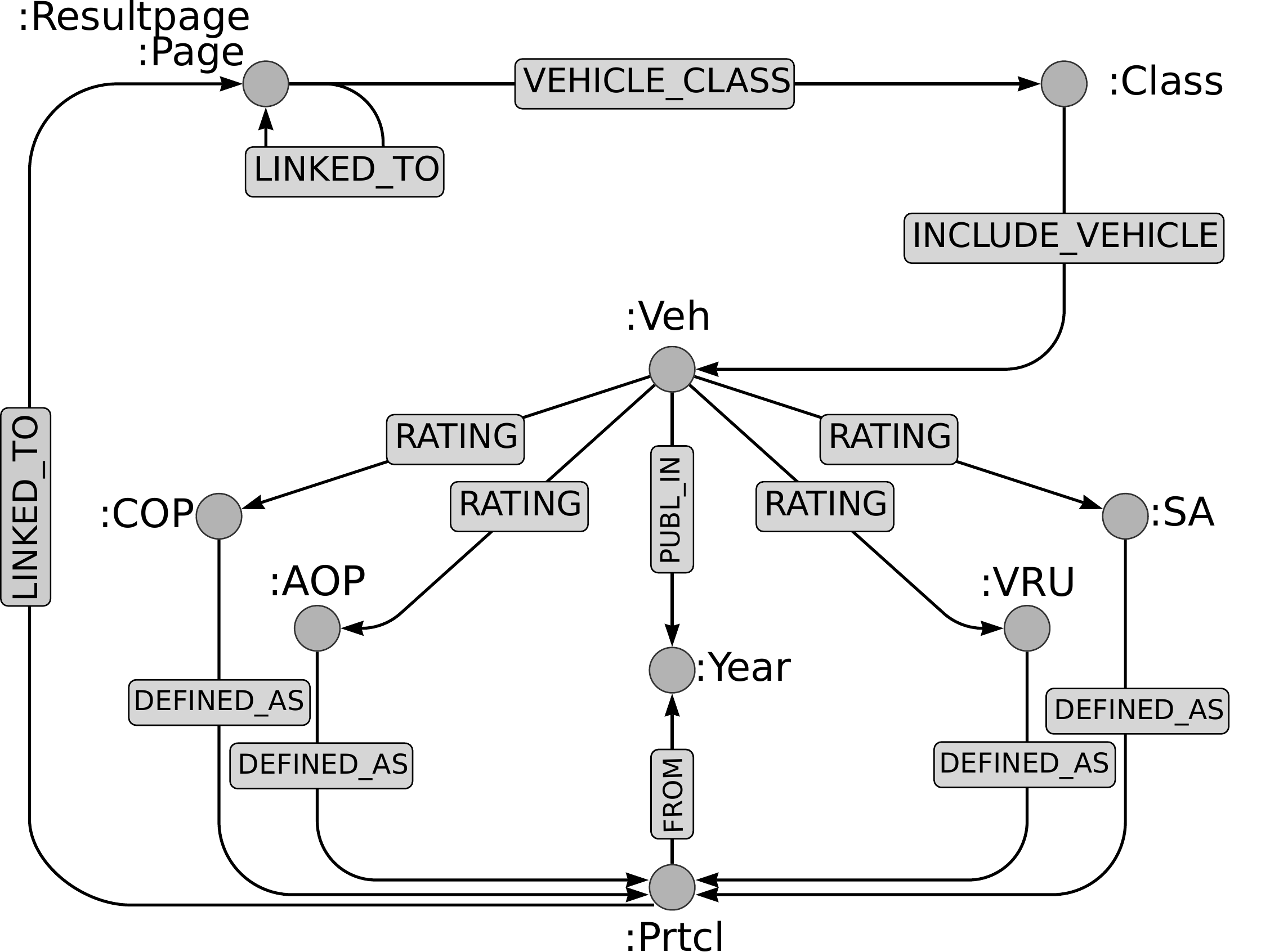}
    \caption{Euro NCAP graph schema.}
     \label{fig:En_schema}
\end{figure}

Furthermore, we model the four test subdisciplines of \textbf{V}ulnerable \textbf{R}oad \textbf{U}ser/pedestrian, \textbf{A}dult \textbf{O}ccupant \textbf{P}rotection, \textbf{C}hild \textbf{O}ccupant \textbf{P}rotection, and \textbf{S}afety \textbf{A}ssist with \mybox{VRU}, \mybox{AOP}, \mybox{COP}, and \mybox{SA} nodes, respectively, Figure~\ref{fig:En_schema}.
Additionally, for each pair of subdiscipline and vehicle, the performance value of a vehicle is stored as the weight of a \myedge{}{RATING}{}{1} edge.
We can extract this information from each tab of the result table, Figure~\ref{fig:result_page}(a).
The specifications for subdisciplines may differ for each vehicle depending on the market release year of the vehicle.
Therefore, we add a \mybox{Year} node connected to each \mybox{Veh} to identify the relevant protocols for each vehicle.

Generally, we aim to have a continuous data extraction and import from the Euro NCAP pages.
To achieve that, we additionally store data about the webpage, where \mybox{Page} refers to each page as a node in the database, and page connectivity, \myedge{}{LINKED\_TO}{}{1.3}, reveals the webpage's structure and its content.
The URLs of protocols available on the Euro NCAP pages are structured based on the year and the subdiscipline.
We use this structure to generate an additional node \mybox{Prtcl} for each protocol URL and connect the protocol PDF to its year and subdiscipline with \myedge{}{DEFINED\_AS}{}{1.3}.

%
%
\section{CAE Data Model}
\label{sec:cae_data}

\begin{figure*}
    \centering
      \includegraphics[width=0.95\linewidth]{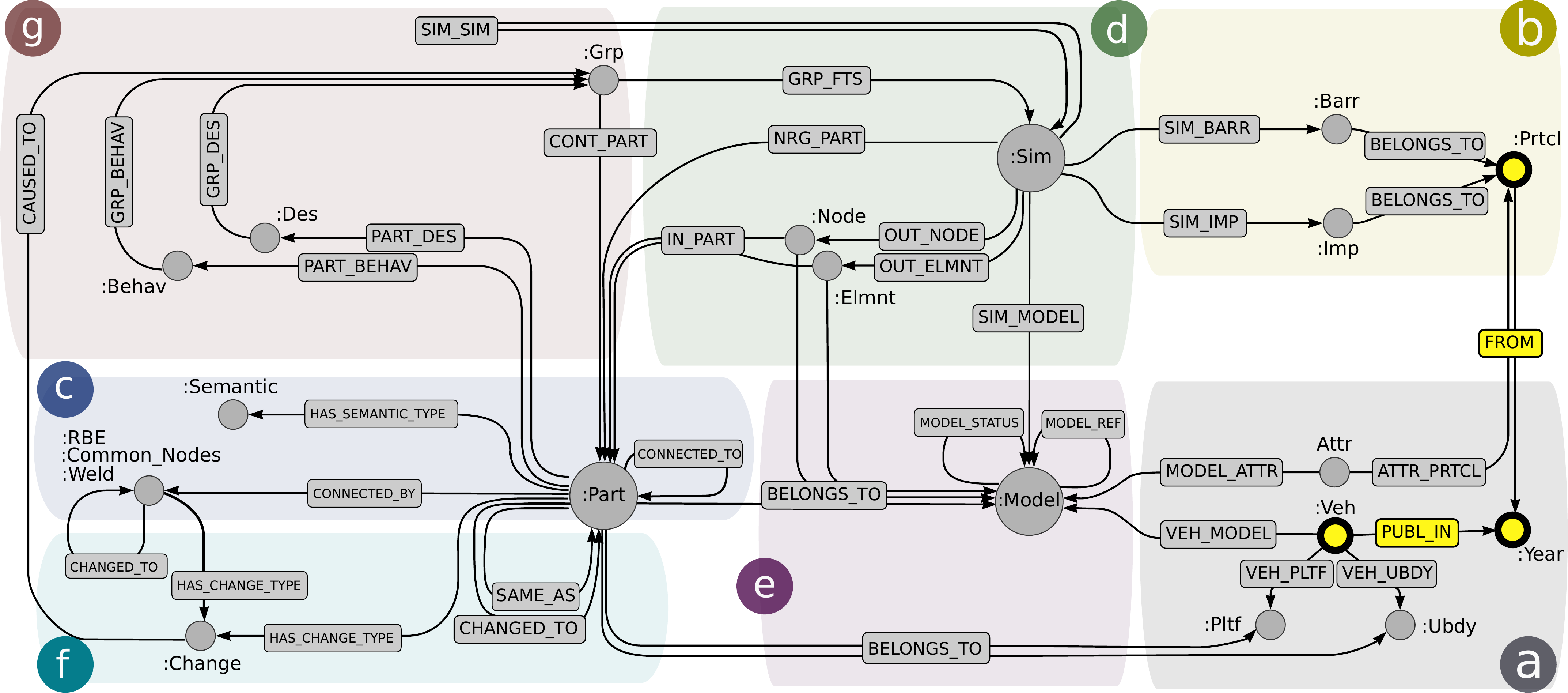}
      \caption{CAE data graph schema, the lower parts reflect the overall development process, and the upper the analysis discipline. (a) portion of the vehicle development process, (b) safety requirements for a vehicle, (c) FE-model input, (d) FE-simulation result, (e) development history of the analysis, (f) FE-models comparison, and (g) ML applications. }
    \label{fig:CAE_schema}
\end{figure*}

For the graph modeling of CAE data, we choose the graph nodes to represent the hierarchy of a vehicle development process down to the FE representations.
Our graph modeling consists of several segments, which are split based on the data sources or data processing, Figure~\ref{fig:CAE_schema}(a) to (g).
It starts with a high level representation of the development structure relevant to CAE data, as well as the vehicle safety regulations, Figure~\ref{fig:CAE_schema}(a) and (b), respectively.
The combinations of both describes the setup and input for the CAE analysis.
The other segments,~\ref{fig:CAE_schema}(c) to (g), are more specific to CAE data.

For the CAE segments, the two main stages differentiate between the data from the input and output of the CAE analysis, Figure~\ref{fig:CAE_schema}(c) and (d), respectively.
The evolving analysis over time connects the CAE data to the development of the vehicle structure, where we dedicate a separate segment to this, Figure~\ref{fig:CAE_schema}(e), bridging (a) and (c).
Finally, the last two segments contain the additional semantics on top of the raw data for input and output, Figure~\ref{fig:CAE_schema}(f) and (g), respectively.
These two build the basis for and represent the further analysis outcome of the car-graph, e.g. for assisting in processing the results by comparing simulations, analysis aiming to connect cause and effect, or methods for the recommendation of solutions.
In the following, we give further details of each segment and present most of the proposed semantics.

We start with the top level, the final output of the company as a vehicle, and go down to the smallest entity of the FE analysis, nodes and elements.
A vehicle is typically divided into the so-called platform and upper body.
One broad definition of a platform is a relatively large set of product components that are physically connected as a stable sub-assembly and are common for different final vehicles~\cite{meyer1997power}.
Using a platform approach, a company can develop a set of differentiated products~\cite{wheelwright1992revolutionizing}.
For simultaneous developments, what deviates between the vehicles, e.g. sedan versus minivan, is the upper body.
We found this concept essential to be captured in the data modeling, since it enables the comparison of related vehicles, i.e. different upper bodies using the same platform.
As a result, a vehicle \mybox{Veh} node has an upper-body \mybox{Ubdy} and a platform \mybox{Pltf}, Figure~\ref{fig:CAE_schema}(a).

As the next segment, let us consider the safety requirements.
Part of the vehicle development process is fulfilling crash test protocols, where a third party usually defines the requirements for crash safety, e.g., Euro NCAP. Note that a crash analysis always includes a vehicle and a type of test, specified by a barrier or an impactor. During the car development process, a set of barriers or impactors are studied based on pre-defined crash test scenarios.
Consequently, connecting each vehicle to crash test protocols and the barrier or impactor provides an overview of the type of analysis included in the CAE-data, where the yellow nodes on Figure~\ref{fig:CAE_schema} connect to the Euro NCAP schema from Figure~\ref{fig:En_schema}, see also section~\ref{sec:en_data}.

Usually, the FE-model representation of barriers or impactors is not changing during vehicle development, while there are slight deviations such as positioning or mass for the robustness studies.
However, technically the vehicle and, say, the barrier are in one FE-model; we separate the barrier from the vehicle in the proposed CAE data model.
This way, different crash scenarios share the same FE-model from the vehicle design.
With such a separation, we enable the reuse of the input FE-model for different analyses, easy reporting of the complete safety status of a single design, and reducing computational time for comparing the inputs by disregarding the changes due to barriers or impactors.
Consequently, an FE simulation \mybox{Sim} is the combination of a \mybox{Model} representing the designed vehicle and barrier \mybox{Barr} or impactor \mybox{Imp}.

The input and output of the solver compose the main share of CAE data, dividing the data into the FE-model and simulation level.
The FE-model level represents the configuration at the start of the simulation, Figure~\ref{fig:CAE_schema}(c).
At the simulation level, deformations, and other mechanical properties over time, quantify the simulated behavior, Figure~\ref{fig:CAE_schema}(d).
Additional metadata can be extracted from the input configuration of the simulation software, these describe the crash scenario, the vehicle characteristics, the development state, or the model/simulation parent.
This parent-child relationship of simulations shapes the so-called development tree of a vehicle, \myedge{}{MODEL\_REF}{}{1.3}, and provides the possibility to reuse former results for the missing result via \myedge{}{MODEL\_STATUS}{}{1.55} connectivity, Figure~\ref{fig:CAE_schema}(e).
Here, each FE-model \mybox{Model} has an edge to the \mybox{Veh} node and several models exist for a vehicle configuration during the development.

For filling the CAE data model with actual data, a challenge is to convert a given \mybox{Model} or \mybox{Sim} to the corresponding graph representation.
Formally, the number of keyword entities in a single simulation is almost 4000 times more than the total number of available simulations for a developed vehicle.
Consequently, we utilize domain knowledge in data processing and define corresponding semantics for the import, to leverage the benefits of the application of KGs for CAE data.

First, \mybox{Model} connects to all entities in the FE-model.
In modeling the FE-model, we examine semantics to classify changes, with the future goal to generate and recommend new models based on the development tree.
We keep the FE-model semantic resolution on the part level and their connectivity in this state.
Roughly speaking, parts in an FE-model are a group of elements with the same properties, e.g., thickness and material.
Each part of an FE-model resembles a node in our graph and has connections to its neighboring parts based on its connectivity type, Figure~\ref{fig:CAE_schema}(c).
Furthermore, we add an edge from parts to \mybox{Pltf} or \mybox{Ubdy} to capture the structural role of the part.

The specific FE-model entity selection strategy follows the result of the ModelCompare software~\cite{garcke2017modelcompare}.
ModelCompare compares two similarly discretized finite element models and organises their differences in a semantic fashion.
Its outcome contains the pairwise comparison of the models and summarizes the changes.
Accordingly, \mybox{Change} is semantic for the changes of compared models.
With this concept, it could be sufficient to model each \mybox{Part} to the database only if there is a change in the part.
However, this way of modeling is not appropriate for the simulation data.
Deformation of the same designed part differs due to its neighboring part, as long as slight changes are applied.
Consequently, changes detected by ModelCompare are stored as \myedge{}{CHANGED\_TO}{}{1.5} edges, whereas all remaining, unchanged parts are connected via \myedge{}{SAME\_AS}{}{1.1} for the compared models, Figure~\ref{fig:CAE_schema}(f).
Finally, note that \mybox{Semantic} nodes are generated as a parts container. This enables us to capture the case of a part being split into several parts from one FE model to the next.

Second, for the data modeling of simulations, a \mybox{Sim} node reflects a FE simulation outcome, where its properties stem from global entities of the simulation, e.g., total mass or impact energy.
Similar to the FE-model data, parts are the main entities representing the simulation, \mybox{Part} Figure~\ref{fig:CAE_schema}(d).
However, this connection reflects the CAE analysis characteristic and may differ based on the type of the analysis.
For crash simulations, the core problem is assessing the impact's energy absorption and managing the energy flow to prevent human injuries and energy features summarize the development fingerprint~\cite{pakiman2022} .
Consequently, as an important outcome information, we use certain energy features of the parts to connect them via \myedge{}{NRG\_PART}{}{1.3} relationship to the \mybox{Sim}.
Note that thereby \mybox{Part} contains properties and relations from both the FE-modeling and simulation level, e.g., thickness and energy absorption features, respectively.

The highest resolution level of a simulation is node and elements data. These entities' deformations and mechanical properties are outputs in each time step.
Considering all these details may miss, say, small critical deformations overlayed by the overall deformation, while analysing only at the part level is too coarse.
Therefore, in the CAE modeling setup, some nodes and elements are defined as output sources for engineers, e.g., elements for cross-section output or nodes for accelerometer or intrusion assessment.
For now, we limit our storage to these selected nodes and elements, \mybox{Node} and \mybox{Elmnt} respectively, Figure~\ref{fig:CAE_schema}(d).

Note that our eventual analysis aims are to detect simulation outliers, assess similarities of simulations, predict the so-called crash load-path, and provide a rough prediction of properties from one vehicle to another, in particular those sharing a platform design.
ML analyses mainly extract features from the simulations or FE-models to assess similarities, e.g., feature engineering or dimensionality reduction.
Extracted features depend on the analysis grouping setting, e.g., group of parts or simulations.
To store outcome of these analysis we introduce three additional nodes \mybox{Des}, \mybox{Behav}, and \mybox{Grp} Figure~\ref{fig:CAE_schema}(g).
Design \mybox{Des} and behavior \mybox{Behav} bundle parts with similar features at FE-model and simulation, respectively.
\mybox{Grp} is a group of different graph nodes in the database.

Several scenarios exist for \mybox{Des} and \mybox{Behav} connectivity.
First, they connect to the stand-alone entities that address the entity-based bundling between many simulations.
In this case, the features are independent of the grouped study.
Furthermore, performing the analysis for a group of entities is sometimes beneficial, e.g., parts as a component or simulations as a development tree.
A component is a group of parts whose role in the crash analysis is the same.
In this case, the \mybox{Des} and \mybox{Behav} are connected to the \mybox{Grp} instead and contain the information about the grouped entities.

Finally, we add \myedge{Grp}{GRP\_FTS}{Sim}{1.3} 
as a weighted edge to store feature extraction for grouped entities in the simulation level in addition to the possibility of direct aggregation of features extracted per part.
To summarize, if \mybox{Des} and \mybox{Behav} connect to the FE-model or simulation entities directly, they are entity dependent, and if they connect to \mybox{Grp}, they are group dependent.
These three nodes can store the ML analysis output on any CAE data e.g., \mybox{Node}, \mybox{Elmnt}, \mybox{Sim}, \mybox{Model}.

Finally, we observe that it is common to share the FE-model between different disciplines during vehicle development.
Our graph modeling's initial target is crash safety, as the most complicated analysis in model size and computational time compared to other solid mechanics FE analyses.
However, our graph modeling is still applicable for other solid mechanics CAE domains, with corresponding semantics for the requirements and the output quantities to assess the simulation, segment (b) and (d) respectively.
To enable this extension, we add an \mybox{Attr} to identify the analysis discipline with its attribute, e.g., safety and durability, in Figure~\ref{fig:CAE_schema}(a).

%
%
 \section{Applications}
\label{sec:applctn}

In this work, we introduce an initial graph for crash simulations and Euro NCAP safety assessments as an example of CAE data and its requirements.
The graph modeling enables two main applications, dynamic semantic-oriented reporting of the data and ML assistance in analysing the results.
Consequently, we support the development and uptake of car-graph with a web-based platform called CAEWebVis\footnote{Accessible at \href{https://CAEWebVis.scai.fraunhofer.de/}{CAEWebVis.scai.fraunhofer.de/}.} to enable semantic reporting for CAE.

A web-based reporting for CAE is a direct usage of the data.
Web-based platforms for CAE data were introduced as early as CAE-bench~\cite{Hagele2000} and recently as an open-source CAE platform~\cite{Lee2020}.
Still, even after two decades, documentation using web technologies is not sufficiently integrated at most OEMs.
We identify as obstacles the high overhead for implementation, lack of software independence from the company data structure, and usage of relational databases, which are not as agile as graph databases.
We employ a web-based user interface that enables project members from different teams to access the CAE results by utilizing the advantages of web standards for ease of use, in particular without detailed CAE experience.
Consequently, we propose a framework that auto-generates and organizes results at a middle level and presents it as a web page on the company server.
CAEWebVis envisions a data organization following web semantics, where it connects simulation results to attribute requirements and design limitations.
Such a connection of documents will increase the learning rate between the different disciplines.
Ultimately, these semantics enable the establishment of car-graph.
CAEWebVis's ultimate vision is to have a tool with low overhead to support the developments required to build safer cars.

The developed CAEWebVis framework for an enhanced exploration of the data is based on our graph modeling.
We use component-based development to have similar visualization of data for different attributes and analyses.
The platform's target is to link data from requirements to assess CAE performance and competitive market performance.
The concept of graph modeling combined with the micro-component web development concept provides enormous flexibility and efficiency in reporting.
This flexibility is an outstanding capability since changes to CAE data are often, e.g., improvement of an impactor modeling or a new crash scenario with a bicycle.
Moreover, ML methods are not yet established in this domain, which needs flexibility in data modeling.
Additionally, the newly provided flexible way of reporting empowers users to interactively and efficiently pick and combine different visualizations and dynamically compare analysis among many simulations.
This is to be seen in contrast with the traditional way, where it is time-consuming to collect different reports from several engineers and one typically for a single design only has a statistical overview of what differs among several hundred simulations.

As an example let us consider pedestrian safety requirements, these assess the vehicle's safety for four types of impactors: lower leg, upper leg, and child and adult head impact. 
For each impactor, a range of probable impact positions are employed. 
Consequently, each vehicle design leads to about 250-350 simulations, which highlights the benefits of dynamic reporting.
Here, consider the assessment measures for head impact and upper and lower leg.
The connection of this visualization to the protocols, e.g., Euro NCAP, supports the CAD engineer in a better understanding of the problem and assures attribute leaders that the project is following the correct variant of assessment.
Additionally, the usage of the introduced representation of different data sources in one graph model enables a quick comparison of the vehicle under development with the market performance.

To achieve this, CAEWebVis has two types of views: zoom-in and zoom-out views.
These views maintain classical reports and ML outcomes.
For the zoom-in view we use visualizations that are simply assessable for humans, e.g., a single in-detail visualization such as a one-pager with animations and curve views, e.g., for the accelerometer or section forces, or combined view for less than ten simulations.
On the other hand, the zoom-out view is for looking at many simulations simultaneously, such as a scatter-plot view for ML embedding results or a summary status-view for protocols assessment for many simulations.
Based on the proposed graph modeling, these visualizations can be agile and efficiently adapt for different crash scenarios, ML analysis, and other CAE data in broader aspects, e.g., visualizing the status of a model, entities in segments (d) and (g) in Figure~\ref{fig:CAE_schema}, based on its connectivity to different barriers and attributes.

Regarding ML applications, we have earlier methods~\cite{IzaTeran.Garcke:2019}, and ongoing research~\cite{Steffes-lai.ea:2021,pakiman2022}, that we are transferring to exploit the introduced data model.
Furthermore, one of the properties often used in graph mining is the node's degree and corresponding ranking of the nodes.
For example, following a high degree \mybox{Change} can be a good recommendation for inexperienced engineers.
Moreover, \mybox{Des} and \mybox{Behav} nodes with a high degree reflect common parts and CAE analysis features, respectively, in a development stage, which highlights fundamental parts and essential times during the crash.
On the other hand, low degree \mybox{Des} nodes reflect components that are outliers or in unexplored design space.
Additionally, cross-domain parts are easily identified with querying nodes with high-degree changes that are common between attributes.

Nevertheless, using graph algorithms to gain insights from CAE data is still challenging.
Potential analysis goals include predicting the similarity of simulations via \myedge{}{SIM\_SIM}{}{1.1} or cause and effect predictions for each simulation \myedge{}{CAUSED\_TO}{}{1.3}.
These are weighted edges, where the weight refers to the similarity predictions between the simulations and the strength of linkage of the cause and effect from model changes to the simulation behavior, respectively.
In the past years, the research on graph-algorithms has focused on methods for analysing data such as social media interactions or fraud detection, where the graph is homogeneous.
For CAE data, a significant obstacle is to fit the data modeling to the current capability of the available ML methods.
For example, in~\cite{pakiman2022}, we used Simrank++ to assess the simulations' similarity and used graph visualization techniques to summarize all analysed simulations in different development phases.
However, a change in the data modeling required merging two nodes' information to generate a bipartite graph to make existing methods applicable.

In addition to the difference in the data entities and interactions, CAE data has in comparison also different proportions of data size.
Recently GNN methods have been influential in graph analysis. However, the data size typically required for these methods is not fulfilled in the CAE domain.

%
%
\section{Conclusions and outlook}
\label{sec:conclusion_and_outlook}

The complexity of CAE raw data and the lack of semantics in the current vehicle development workflow causes design engineers and attribute leaders to rely on CAE engineer reporting.
However, this static reporting restricts the independent exploration of the data.
The lack of semantics in CAE data makes the data disconnected and hinders multi-disciplinary collaboration, which degrades efficient problem-solving.
We envision that the introduced car-graph empowers semantic reporting for CAE.
It should enable project members from different teams to access the CAE results, understand the design performance limitations, compare simulations, and use ML algorithms on the car-graph.

We introduced initial data modeling for CAE data, which enables searchability, filtering, recommendation, and prediction during the development process.
Besides, we linked CAE data to the required protocol and considered comparing vehicle safety performances.
The presented graph modeling uses CAE crash simulation as a source for the vehicle development process and the Euro NCAP webpage as the source for assessment and benchmarking.
Consequently, we developed the platform CAEWebVis as an example of semantic reporting based on graph modeling to illustrate to automotive engineers the advantage of dynamic reporting.
The engagement and feedback of OEMs shall enrich the semantics, which would empower car-graph in exploring and predicting CAE data and step toward safer vehicles.
However, it is still the early stage of car-graph research, and further research is ongoing for empowering ML algorithms on CAE data that will extend this graph modeling.

Introducing the car-graph problem to the KGs community promotes a new type of problem for graph analytics.
Note that the difference in problem type results in a shortage of available analysis packages.
NetworkX~\cite{NetworkX} and DeepSNAP~\cite{DeepSNAP} are examples of Python libraries to assist efficient analytics and deep learning on graphs.
The usage of NetworkX is hampered since the effects of edge weights differs in the CAE domain in contrast to, say, analysing customers.
Furthermore, DeepSNAP methods are available for a heterogeneous graph, and its methods look primarily at having multiple nodes for labeling.
However, in engineering, having multiple edges implies having different functions in the massage passing, for which modified or new methods or needed.
We hope that the automotive industry's needs will increase the interest in research on different types of graph algorithms in the future.

Regarding the CAE data modeling, there is further information available that is beneficial to include and will further enhance the graph model.
For example, we aim to include more semantics of the configuration of a vehicle, e.g., engine or roof.
Moreover, incorporating CAD data such as component definitions, packaging, design guidelines, and manufacturing limitations on the design would bring the CAE engineers to the next level for efficient problem-solving.

Furthermore, Euro NCAP is just one of many safety assessment protocols, and other protocols and requirements are missing.
For example, a step can be to extract structure from the existing PDFs of the requirements, or, ideally, work with Euro NCAP to directly provide the corresponding information with semantics.
Additionally, many vehicle characteristics such as vehicle dimension or weight distribution, while unavailable from Euro NCAP, exist in other public data sources and can enrich the data.

Current graph modeling's conclusive path is the cause, effect, and safety assessment.
The outlook of this path will be recommendations for better design solutions during the development process and improvements of multi-disciplinary design guidelines between projects, enabled by a knowledge graph for crashworthiness.
Furthermore, future vehicle safety aspects consider assisting active safety systems with passive safety characteristics, as well as supporting broader crash scenarios and diversity in crash analysis.
Therefore, we envision the car-graph bridging autonomous driving to product identity during the development with CAD databases to include vehicle configurations.
In this way, the car-graph shall allow an extension of the safety evaluation from regulated tests, which are just examples of real crash scenarios, to more diverse crash scenarios.
As a result, car-graph will couple active and passive safety to associate autonomous driving with safe crashes in situations where a crash is inevitable.

%
%
\bibliographystyle{ieeetr}
\bibliography{cargraph_ontology.bbl}

\end{document}